\documentclass[runningheads]{llncs}
\RequirePackage{amsmath}
\usepackage{graphicx}
\usepackage[utf8]{inputenc} 
\usepackage[T1]{fontenc}   
\usepackage{url}           
\usepackage{booktabs}     
\usepackage{amsfonts}
\usepackage{nicefrac}      
\usepackage{microtype}      
\usepackage{enumerate}
\usepackage{graphicx}
\usepackage[caption=false]{subfig}
\usepackage{algorithmic}
\usepackage{algorithm2e}

\begin{document}
\title{
On Calibrated Model Uncertainty in Deep Learning}

\titlerunning{Calibrated Model Uncertainty in Deep Learning}
%
%

\author{Biraja Ghoshal\inst{1}
\and
Allan Tucker\inst{1}
}

\authorrunning{B. Ghoshal et al.}
%
\institute{Dept of Computer Science, Brunel University, \\London, Uxbridge, UB8 3PH, United Kingdom 
\\
\email{biraja.ghoshal@brunel.ac.uk}\\
\url{https://www.brunel.ac.uk/computer-science}}
\maketitle              

\begin{abstract}
Estimated uncertainty by approximate posteriors in Bayesian neural networks are prone to miscalibration, which leads to overconfident predictions in critical tasks that have a clear asymmetric cost or significant losses. Here, we extend the “approximate inference for the loss-calibrated Bayesian framework” to dropweights based Bayesian neural networks by maximising expected utility over a model posterior to calibrate uncertainty in deep learning.  Furthermore, we show that decisions informed by loss-calibrated uncertainty can improve diagnostic performance to a greater extent than straightforward alternatives. 
We propose Maximum Uncertainty Calibration Error (MUCE) as a metric to measure calibrated confidence, in addition to its prediction especially for high-risk applications, where the goal is to minimise the worst-case deviation between error and estimated uncertainty. In experiments, we show the correlation between error in prediction and estimated uncertainty by interpreting Wasserstein distance as the accuracy of prediction.  
We evaluated the effectiveness of our approach to detecting Covid-19 from X-Ray images. Experimental results show that our method  reduces miscalibration considerably, without impacting the model’s accuracy and improves reliability of computer-based diagnostics.
\end{abstract}


\section{Introduction}
\label{submission}
Recently, Deep Learning has achieved state-of-the-art performance across applied sciences, engineering and medical imaging. Neural networks trained by minimising a cross-entropy loss tend to overfit based on classification accuracy. Approximate inferences such as Variational inference \cite{blundell2015weight,YarinGal2016Thesis,ghoshal2019,graves2011practical} and Markov chain Monte Carlo (MCMC) \cite{neal2012bayesian} which approximates the posterior distribution in Bayesian neural networks (BNN), are prone to miscalibration, and the estimated uncertainty does not always represent the error in model prediction. This can lead to overconfident predictions, which raise concerns over its safety in critical applications \cite{cobb2018loss}. In practical decision-making systems, the cost of falsely misdiagnosing a disease when a patient is infected (i.e. a false negative) may be much more significant, than incorrectly diagnosing a disease when it is not present (false positive). It is imperative that all Bayesian Deep Learning models should provide a calibrated uncertainty measure in addition to its point prediction, especially when the probability associated with the predicted disease is low. 

Coronavirus (COVID-19) represents a new strain of Coronavirus and presumably a mutation of other Coronaviruses \cite{shan+2020lung}. Dealing with it is a currently a significant medical challenge around the world. The existing dataset, which consists of limited image data sources with expert labelled data set, for the detection of COVID-19 positive patients is insufficient, and manual detection is time-consuming. Our goal is to provide reliable Deep Learning based solution, combined with clinical practices to provide automated detection with estimated bias-reduced well-calibrated uncertainty to aid the screening process. 

Bayesian decision theory combines uncertainty with task-specific utility function \(U(\theta, a)\), which extends the Bayesian paradigm, the utility of taking action \(a \in \mathcal{A}\). This maximises the expected utility over the posterior to make rational predictions in state \(\theta\). The overall process is computed using a 2-step procedure: probabilistic inference and optimal prediction. First, approximate the posterior \(p(\theta|D)\) with a \(q(\theta|D)\) and then minimise 
evidence lower bound (ELBO) loss that incorporates the network weights and task-specific utility function under \(q\), where we assume that approximate \(q\) measures properties of the posterior. A clearly defined goal of a prediction is necessary as an evaluation criterion in the form of a utility function. Therefore, this should jointly optimise the approximate posterior with the action that maximises the expected utility with respect to the posterior over the model parameters, which will minimise the posterior risk. 

Bayesian Neural networks, provide a probabilistic interpretation of deep learning models and a principled method for modelling uncertainty. However, estimated model uncertainty in deep learning often fail to capture the true distribution of data and so is prone to miscalibration. As a result, calibrated model uncertainty is essential to ensure that uncertainty measure from the model is useful as confidence is crucial in safety-critical applications. Neural networks with a weighted cross-entropy loss can lead to model over-fitting \cite{cobb2018loss}. In practice, several non-parametric and parametric calibration approaches such as isotonic regression, Platt scaling, temperature scaling (TS) \cite{guo2017calibration} or parametric multiclass Dirichlet calibration have been extensively studied in neural networks \cite{kull2019beyond}.

Cobb et. al. \cite{cobb2018loss} showed that minimising the KL divergence between an approximate posterior \(q\) and a calibrated posterior scaled by the utility function results in the standard evidence lower bound (ELBO) loss for Bayesian neural network inference, as well as an additional task-specific utility function, dependent regularisation term, to stochastic optimisation. This can be implemented as a novel penalty term to the standard neural network.

Following Gal \cite{YarinGal2016Thesis}, Ghoshal et al \cite{ghoshal2019estimating}  showed that Neural Networks with dropweights applied in the fully connected layer, is equivalent to variational Bayesian neural networks. "Dropweights" which randomly drops connections, where weights in the neural networks are set to zeros during both training and inference robust to over-fitting, and can be seen as a form of regularisation \cite{ghoshal2019journal}.

We extended the classic technique to ‘approximate inference for the loss-calibrated Bayesian framework’ \cite{lacoste2011approximate,cobb2018loss} for dropweights based Bayesian neural networks, and so obtained well-calibrated model uncertainty.

Given \(\hat{y}\) is the class prediction of the model and \(\hat{p}\) is its associated confidence, a model is calibrated only if confidence in a prediction matches its probability of correctness \cite{wenger2019non}:
\begin{equation}
    {E}\left [{1_{\hat{y} = y} \mid \hat{p}}\right ] = \hat{p},
\end{equation}

Expected Calibration Error (ECE) is a well-known measure of the degree of calibration to quantify the miscalibration of the difference in expectation between confidence and accuracy \cite{guo2017calibration,naeini2015obtaining,wenger2019non}.
\begin{equation}
    \text{ECE}_q = {E} [ | \hat{p} - {E} [1_{\hat{y} = y} \mid \hat{p} ] |^q ]^{\frac{1}{q}}
\end{equation}
for \(1 \leq q < \infty\). The maximum calibration error (MCE) can be defined as \cite{wenger2019non}
\begin{equation}
    \text{MCE}_{\infty} = \max_{p \in [0,1]} | p - {E} [1_{\hat{y} = y} \mid \hat{p} = p ] |.
\end{equation}

David et. al \cite{ruhe2019bayesian} derived analytically bounds on cross-entropy loss with respect to predictive uncertainty and showed that uncertainty can mitigate performance risk and loss. We expect better calibrated uncertainty in variational Bayesian inference by using loss-calibrated Bayesian deep learning model.  We demonstrate the effectiveness of our approach in detecting Covid-19 from X-Ray images. We show that the estimated uncertainty in predictions have a strong correlation with classification error in prediction. 

\section{Loss-Calibrated Approximate Bayesian Inference Method}

A Bayesian Neural Network (BNN) is a neural network with a prior distribution on its weights, which is robust to over-fitting (i.e. regularisation). Exact inference is analytically intractable, and hence the approximate inference has been applied instead.

Given \({D} = \left\{{X}^{(i)}, Y^{(i)}\right\}\) where \(X \in R^{d}\) is a d-dimensional input vector and \(Y \in\{1 \ldots C\}\), given C class label, a set of independent and identically distributed (i.i.d.) training samples size \(N\), a BNN is defined in terms of a prior \(p(w)\) on the weights, as well as the likelihood \(p(D|w)\). Variational Bayesian methods approximate the true posterior by maximising the evidence lower bound (ELBO) between a variational distribution \(q(w|\theta)\) and the true posterior \( p(w|D)\) w.r.t. to \(\theta\). The corresponding optimisation objective or cost function is
\begin{equation}
    \mathcal{F}(\mathcal{D},{\theta}) = 
    {E}_{q({w} \lvert {\theta})} \log p(\mathcal{D} \lvert {w})
-
\mathrm{KL}(q({w} \lvert {\theta}) \mid\mid p({w})) 
\end{equation}
The first term is the expected value of the likelihood w.r.t. the variational distribution and is called the likelihood cost. The second term is the Kullback-Leibler (KL) divergence between the variational distribution \(q(w|\theta)\) and the prior \(p(w)\) and is called the complexity cost. 

Approximate posteriors in Variational Inference (VI) is prone to miscalibrations and is not sufficient for making optimal decisions, due to inexact posterior predictive distributions. While still retaining a reasonable posterior approximation, instead of maximising the approximation accuracy, Lacoste-Julien et al. \cite{lacoste2011approximate} proposed a loss-calibrated approximate inference, to maximise the expected utility computed over the approximating distribution. 

Given a posterior distribution \(p(\theta|D)\) on data \(D\), an optimal decision \(h\) and utility \(u(\theta, h) >= 0\) defined over the parameter \(\theta\), maximises the posterior gain or alternatively, utility maximisation

\begin{equation}
\mathcal{F}_u(h) = \int p(\theta | D) u(\theta, h)d{\theta}
\end{equation}
However \(p(\theta|D)\) is intractable. 

Variational inference approximate the posterior \(p(\theta|D)\) with \(q _\lambda (\theta)\) parameterized by \(\lambda\) typically by maximizing a lower bound \(L_{VI}(\lambda)\) for the marginal log-likelihood

\begin{equation}
\log p(D) > 
\int{q_{\lambda}(\theta) \log 
    {{P(D|\theta)} \over {q_{\lambda}(\theta)}} d{\theta}} :=  L_{VI}(\lambda)
\end{equation}

Following Lacoste-Julien et al. \cite{lacoste2011approximate} to calibrate variational approximation based on lower bounding the logarithmic gain using Jensen’s inequality as:
\begin{equation}
\log \mathcal{F}_u(h) > L_{VI}(\lambda) + 
{E}_{q}[\log \int (p(y|\theta, D) \mu(h, y)dy] := LC_{VI}(\lambda, h)
\end{equation}

The first term is analogous to the standard variational approximation to provide the final bound. The utility-dependent second term accounts for decision making. It is independent of the observed \(y\) and only depends on the current approximation \(q_\lambda(\theta)\), favouring approximations that optimise the utility.

The Bayesian decision problems formulated in terms of maximising gain defined by a utility \(u(y, h)\geq 0\), or in terms of minimizing risk defined by a loss \(l(y, h) \geq 0\) \cite{cobb2018loss}. To calibrate for user-defined loss, we need to convert the loss into a utility by \(u(y, h) = M - l(y, h)\), where \(M \geq sup_{y.h}(l(y.h)\) with the assumption that the utility to only take positive values.

Recently, Gal \cite{cobb2018loss,YarinGal2016Thesis} proved that a gradient-based optimisation procedure on the dropout neural network is equivalent to a specific variational approximation on a Bayesian neural network. Following Gal \cite{YarinGal2016Thesis}, Ghoshal et. al. \cite{ghoshal2019,ghoshal2020estimating}  showed similar results for neural networks with MC-Dropweights. 
The model uncertainty is approximated by averaging stochastic feedforward Monte Carlo (MC) sampling during inference.
At test time, the unseen samples are passed through the network before the Softmax predictions were analysed.

The expectation of \(\hat{y}\) is called the predictive mean \(\mu_{pred}\) of the model. The predictive mean over the MC iterations is then used as the final prediction on the test sample:
\begin{equation}
\mu_{pred} \approx \frac{1}{T} \sum_{t=1}^{T} p(\hat{y}|\hat{x}, {X, Y})
\end{equation}

For each test sample \(\hat{x}\), the class with the largest predictive mean \(\mu_{pred}\) is selected as the output prediction by pre-defined T Monte Carlo sample. 

Estimation of entropy from the finite set of data suffers from a severe downward bias, when the data is under-sampled; even small biases can result in significant inaccuracies when estimating entropy. We leveraged plug-in estimate of entropy and Jackknife resampling method to calculate bias-reduced uncertainty \cite{ghoshal2020estimating}.

This approach addresses the issues with overconfidence and providing well-calibrated quantification of predictive uncertainty. This is because the uncertainty in weight space for asymmetric utility functions, captured by the posterior, is incorporated into the predictive uncertainty, giving us a way to model “when the machine does not know”. 

\section{Measure of Uncertainty Calibration in Deep Learning}

We describe the most prevalent methods to measure miscalibration of estimated uncertainty associated with the classification. Our loss-calibrated BNN model incorporates asymmetric misclassification costs as a utility function, to enable rejection of uncertain predictions, and so in turn minimise the misclassification, resulting in improved performance, that incorporates practical considerations. 

\subsection{Uncertainty Calibration Error (UCE)}

We propose the following modified notion of Eq. (2) for bias-reduced Uncertainty Calibration Error (UCE) as a measure of the degree of calibration, to quantify the miscalibration of the difference in expectation between model error and estimated uncertainty. 

\begin{equation}
    \text{UCE}_q = {E} [ | \hat{H} - {E} [1_{\hat{y} = y} \mid \hat{H} ] |^q ]^{\frac{1}{q}}
\end{equation}

The estimated bias-reduced uncertainty \cite{ghoshal2020estimating} of a neural network is partitioned into M equally-spaced bins (each of size \(1/M\)), and a weighted average of the  bin error and uncertainty difference. Mathematically, we can express this as \cite{laves2019well}:

\begin{equation}
    \text{UCE} = \sum_{m=1}^{M}{\frac{|B_m|}{n}|\text{error}(B_m)-\text{uncertainty}(B_m)|}
\end{equation}, where n is the number of samples and \(B_m\) is the set of indices of samples whose uncertainty falls into the interval.

We propose the following modified notion of Eq. (3) to quantify the maximum uncertainty calibration error (MUCE):
\begin{equation}
    \text{MUCE} = \max_{m \in \{1, \dots, M\}}{|\text{error}(B_m)-\text{uncertainty}(B_m)|}
\end{equation}

As a measure, MUCE is most appropriate for high-risk applications, where the goal is to minimise the worst-case deviations between error and estimated uncertainty. MUCE calculates the maximum calibration uncertainty for the bins. 

In critical applications, it might be necessary to enforce a low MUCE in order to reduce the risk of overconfidence in prediction. A concise way to visualise the degree of calibration of a model is called Uncertainty-Reliability diagrams.

\subsection{Sharpness}
Sharpness refers to the concentration of the predictive distributions. The accuracy of estimated uncertainty in predictive distributions should be evaluated by maximising the sharpness of the subject during calibration. In practice, estimated uncertainty is not sufficient to calculate the useful probability of making a prediction. For example, a perfectly calibrated binary classifier on a balanced classification problem will always return an uncertainty of 50\%, as this is the probability of making a false prediction. 

Therefore, the sharpness \cite{kuleshov2018accurate} of a model is a good measure of how close the confidence estimates are between 0 and 1. We propose measuring sharpness using the variance as:
\begin{equation}
    \text{sharpness}(f) = {Var} [ |\text{error}(B_m)-\text{uncertainty}(B_m)|]
\end{equation}.

\section{Relationship between Error and Uncertainty}
We show the correlation between the error in a prediction and the estimated uncertainty by interpreting Wasserstein distance (WD) as the accuracy of prediction. \cite{arjovsky2017wasserstein,laves2019quantifying}.
The Wasserstein distance for the real data distribution is \(P_r\) and the generated data distribution \(P_g\) is mathematically defined as the greatest lower bound (infimum) for any transport plan (i.e. the cost for the cheapest plan):
\begin{equation}
W(P_r, P_g) = \inf_{\gamma \sim \Pi(P_r, P_g)} {E}_{(x, y) \sim \gamma}[\| x-y \|]
\end{equation}, \(\Pi(P_r, P_g)\) is the set of all possible joint probability distributions \(\gamma(x, y)\), whose marginals are respectively \(P_r\) and \(P_g\). However, equation (5) for the Wasserstein distance is intractable. Using the Kantorovich-Rubinstein duality, \cite{arjovsky2017wasserstein} the calculation can be simplified to

\begin{equation}
W(P_r, P_g) = \frac{1}{K} \sup_{\| f \|_L \leq K} {E}_{x \sim P_r}[f(x)] - {E}_{x \sim P_g}[f(x)]
\end{equation} where the supremum (sup) is the opposite of the infimum (inf); sup is the least upper bound and \(f\) is a Lipschitz continuous functions \(\{ f_w \}_{w \in W}\), parameterised by \(w\) and the K-Lipschitz constraint \(\lvert f(x_1) - f(x_2) \rvert \leq K \lvert x_1 - x_2 \rvert\). The error function can be configured as measuring the 1 - Wasserstein distance between \(P_r\) and \(P_g\).
\begin{equation}
W(P_r, P_g) = \max_{w \in W} {E}_{x \sim P_r}[f_w(x)] - {E}_{z \sim P_r(z)}[f_w(g_\theta(z))]
\end{equation}, where \(g_\theta(z)\) is the locally Lipschitz continuous functions on \((0, \infty)\).

The advantage of Wasserstein distance (WD), is that it can reflect the distance of two non-overlapping or little overlapping distributions.

\section{Dataset}

The novel coronavirus 2019 (COVID-2019), which results in pneumonia at varying severity, has rapidly become a pandemic.
We have selected 68 Posterior-Anterior (PA) X-ray images of lungs with COVID-19 cases from Dr. Joseph Cohen’s Github repository \cite{cohen2020covid}. This repository is constantly updated with images shared by researchers.  We augmented the dataset with normal and pneumonia images from Kaggle’s Chest X-Ray Images. This has produced a total of 5,941 PA chest radiography images across four classes (Normal: 1583, Bacterial Pneumonia: 2786, non-COVID-19 Viral Pneumonia: 1504, and COVID-19: 68). We standardised and resized all images to 224 x 224 pixels.

\section{Experiment}
We used a pre-trained ResNet50V2 model \cite{he2016identity} and acquired data only to fine-tune the original model. We introduced Dropweights followed by a softmax activated layer, which was then applied in the fully connected layer on top of the ResNet50V2 convolutional base, to estimate a meaningful model uncertainty. 

We split the whole dataset into 80\% and 20\% between training and testing sets respectively. Real-time data augmentation was also applied, leveraging Keras ImageDataGenerator during training, to prevent overfitting and enhance the learning capability of the model. The Adam optimiser was used with a learning rate of 1e-5 and a decay factor of 0.2. All our experiments were run for 25 epochs and the batch size was set to 8. 
Dropweights with rates of 0.3 were added to a fully connected layer. We monitored the validation accuracy after every epoch and saved the model with the best accuracy on the validation dataset. During test time, Dropweights were active and Monte Carlo sampling was performed by feeding the input image with MC-samples 25 through the Bayesian Deep Residual Neural Networks.

\section{Utility Function}

In this study, the utility function in table 1, prescribes for fewer false negatives for Covid-19, Normal, Viral Pneumonia and Bacterial cases, relative to the other categories
from the costs of incorrect diagnoses to a task-specific utility function. 

\begin{table}[ht]
\centering
\begin{tabular}{lllll}
\hline
\multicolumn{1}{|l|}{} & \multicolumn{1}{l|}{Normal} & \multicolumn{1}{l|}{Bacterial Pneumonia} & \multicolumn{1}{l|}{Viral Pneumonia} & \multicolumn{1}{l|}{Covid} 
\\
\hline
\multicolumn{1}{|l|}{Normal} & \multicolumn{1}{l|}{2.1} & \multicolumn{1}{l|}{1.2} & \multicolumn{1}{l|}{1.2} & \multicolumn{1}{l|}{1.2} \\ \hline
\multicolumn{1}{|l|}{Bacterial Pneumonia} & \multicolumn{1}{l|}{1.4} & \multicolumn{1}{l|}{2.1} & \multicolumn{1}{l|}{1.4} & \multicolumn{1}{l|}{1.4} \\ \hline
\multicolumn{1}{|l|}{Viral Pneumonia} & \multicolumn{1}{l|}{1.4} & \multicolumn{1}{l|}{1.4} & \multicolumn{1}{l|}{2.1} & \multicolumn{1}{l|}{1.4} \\ \hline
\multicolumn{1}{|l|}{Covid} & \multicolumn{1}{l|}{1.2} & \multicolumn{1}{l|}{1.2} & \multicolumn{1}{l|}{1.2} & \multicolumn{1}{l|}{2.1} \\ \hline
\end{tabular}
\caption{Maximum utility (2.1) is for correct prediction and the lowest utility (1.2) is given to errors in predicting the Normal and Covid. In safety critical applications, the utility function values to be assigned according to the functional requirements. }
\label{Table1}
\end{table}

\section{Results and Discussions}

\subsection{Model Performance}
The normalised confusion matrices, in Figure 1, demonstrate how the different models compare when making predictions.

\begin{figure}[ht]
    \centering
    \includegraphics[width=12cm]{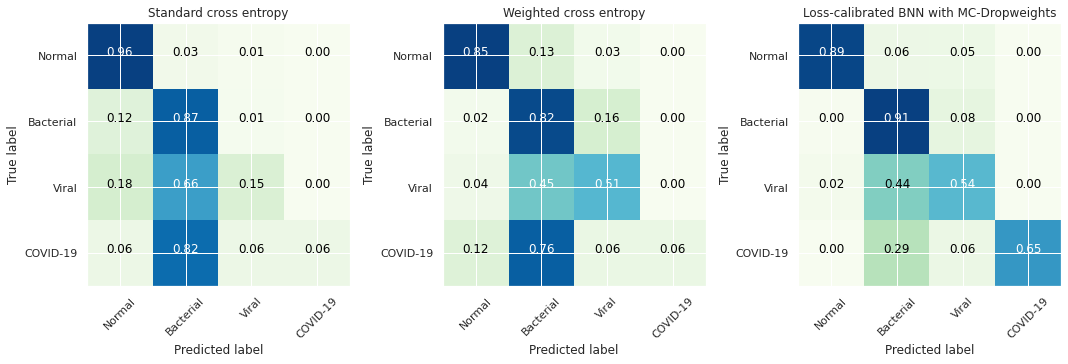}
    \caption{Left: Standard NN model with cross-entry loss. Middle: Standard NN model with weighted cross-entry loss. Right: Loss-Calibrated BNN model. Each confusion matrix displays the resulting classification when averaging the utility function, with respect to the dropweights samples of each network. We highlight that our Loss-Calibrated BNN model captures our preferences, by avoiding false negatives of the ‘Covid-19’ diseased class. There is also a clear performance gain from the loss-calibrated model. This compares favourably to the standard model, where there is a common failure mode of predicting a patient as being ‘Normal’ when they are ‘Covid-19’ infected.}
\end{figure}

\subsection{The Relation between Cost as Expected Loss and Predictive Accuracy}

In this automatic disease detection in X-Ray Images example, our goal is to reduce false negatives, whilst being concerned about false positives. Table 1 below demonstrates that there is a strong correlation between prediction accuracy and the loss as costs of incorrect misdiagnoses to a task-specific utility function, for example the highest cost and therefore lowest utility is assigned for a patient, who is misdiagnosed as being healthy when their condition is severe.

\begin{table}[ht]
\centering
\begin{tabular}{|l|l|l|l|}
\hline
 & Standard BNN  & Weighted Cross Entropy & Loss-Calibrated BNN   \\ \hline
Accuracy (\%): & 70.12 & 74.09  & 80.88 \\ \hline
Expected loss: & 0.20 & 0.16 & 0.12 \\ \hline
ECE: &  13.27 & 4.71 & 7.01  \\ \hline
UCE: &  16.05 & 28.79 & 10.86 \\ \hline
MUCE: & 3.77  & 5.42 & 1.91 \\ \hline
Sharpness: & 0.009 & 0.026 & 0.027 \\ \hline
\end{tabular}

\caption{The result evident from above table 2 is that Loss-Calibrated BNN with MC-Dropweights provides significant improvements in calibrated confidence over measured uncertainty from standard NNs.}
\label{Table 2}

\end{table}
\subsection{Reliability Diagrams}
Reliability diagrams is a visual representation of model calibration \cite{degroot1983comparison,guo2017calibration}. Figure 2 diagrams plot the expected accuracy obtained for each bin (fraction of positives) against the binned predicted confidences. A perfectly calibrated model would result in a 45-degree line. Any deviation from this perfect diagonal represents miscalibration, where a lower ECE (close to zero) indicates a better calibration.

Model Uncertainty-Reliability diagrams in Figure 3 represent the deviation of the perfect calibration by plotting the binned measured model uncertainties against the  error obtained for each bin (fraction of
negatives). The UCE is defined as the absolute error of these bins (i.e., the gap
between uncertainty and accuracy) weighted by the number of samples in the bins, where a higher UCE indicates a better calibration.

\begin{figure}[ht]
    \centering
    \includegraphics[width=4cm]{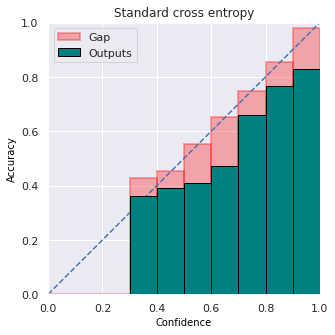}
    \includegraphics[width=4cm]{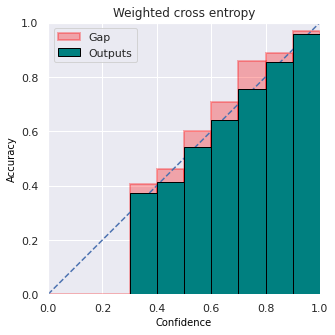}
    \includegraphics[width=4cm]{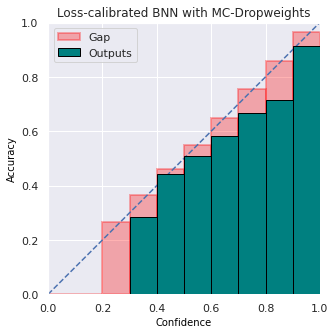}
    \caption{Confidence-Reliability diagrams showing three classifiers (Left: Standard NN model with cross-entry loss. Middle: Standard NN model with weighted cross-entry loss. Right: Loss-Calibrated BNN model.) and its confidence histogram (M = 10 bins) on Covid-19 image dataset.}
\end{figure}

\begin{figure}[ht]
    \centering
    \includegraphics[width=4cm]{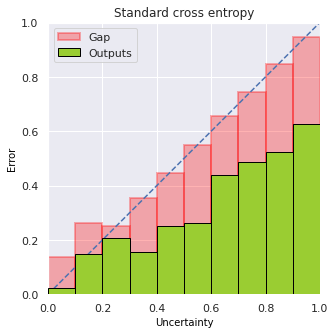}
    \includegraphics[width=4cm]{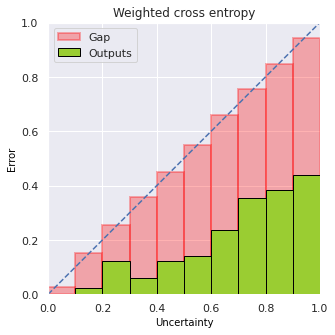}
    \includegraphics[width=4cm]{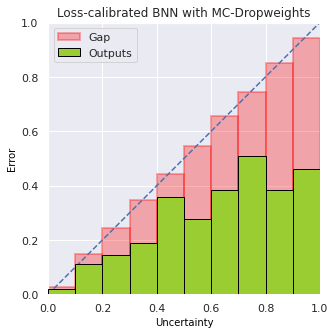}
    \caption{Uncertainty-Reliability diagrams showing two classifiers (Left: Standard NN model with cross-entry loss. Middle: Standard NN model with weighted cross-entry loss. Right: Loss-Calibrated BNN model.) and its Uncertainty histogram (M = 10 bins) for Covid-19 image dataset. Loss-calibration model updates the posterior approximation, so that estimated uncertainty in Bayesian inference for optimal decisions are better in terms of a user-defined asymmetric cost and error in prediction to avoid overconfidence.}
\end{figure}

\subsection{Correlation between Model Uncertainty and Predictive Error}

Figure 4 shows that uncertainty correlates with error in prediction. Well-calibrated deep learning-based model uncertainty is expected to increase safety in critical applications, reducing overconfidence in erroneous prediction with low uncertainty. This will improve trust in deep learning-based detection.

\begin{figure}[!htp]
    \centering
    \includegraphics[width=4cm]{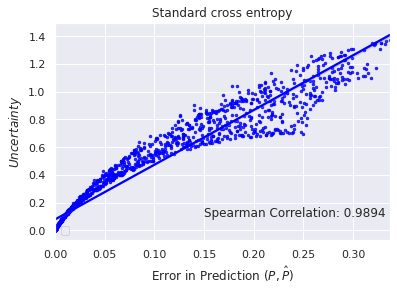}
    \includegraphics[width=4cm]{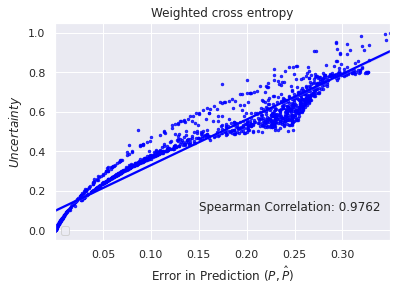}
    \includegraphics[width=4cm]{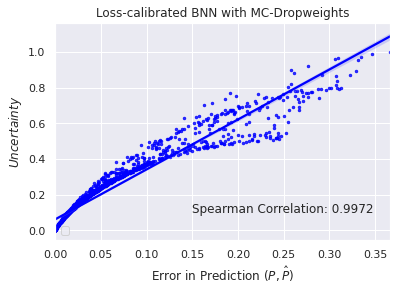}
    \caption{The figure 4 above shows the correlation between estimated bias-reduced uncertainty \cite{ghoshal2020estimating} and the error of prediction with variation of the Wasserstein Distance, thus loss-calibrated model reduces miscalibration.}
\end{figure}

\section{Conclusion and Future work}

Bayesian decision-theoretic loss-calibrated approximate inference calibrates uncertainty obtained in BNN with dropweights achieves encouraging performance when learning an approximate distribution over weight parameters, incorporating uncertainty and user-defined asymmetric utility functions. The significance of our experiment demonstrates the usefulness of loss-calibrated model to large networks with real-world medical imaging Covid-19 diseases detection. Critical decision-making for medical imaging applications requires, not only a high accuracy, but also reliable estimation of predictive uncertainty to interpret the model. Calibrated estimated uncertainty also gives possibilities to identify out of domain data patterns and to reject uncertain predictions so further advances image annotation.

Recent advances in deep learning (e.g. network architecture, optimizer, learning rate, normalisation, regularization, dropweights rate etc.) have effects on calibrated model uncertainty while improving accuracy. It remains future work in general data‐driven Bayesian decision‐making contexts.

%
%
\bibliographystyle{splncs04}
\bibliography{ECML2020-WUML}
\end{document}